\documentclass[10pt,twocolumn,letterpaper]{article}

\usepackage{wacv}              

\usepackage[accsupp]{axessibility}  

\usepackage{graphicx}
\usepackage{amsmath}
\usepackage{amssymb}
\usepackage{booktabs}
\usepackage{xcolor}

\usepackage[pagebackref,breaklinks,colorlinks]{hyperref}
\usepackage{makecell}

\usepackage[capitalize]{cleveref}
\crefname{section}{Sec.}{Secs.}
\Crefname{section}{Section}{Sections}
\Crefname{table}{Table}{Tables}
\crefname{table}{Tab.}{Tabs.}

\def\wacvPaperID{690} 
\def\confName{WACV}
\def\confYear{2024}

\begin{document}

\title{MITFAS: Mutual Information based Temporal Feature Alignment and Sampling for Aerial Video Action Recognition}

\author{Ruiqi Xian$^*$, Xijun Wang$^*$, Dinesh Manocha\\
University of Maryland - College Park\\
{\tt\small \{rxian, xijun, dmanocha\}@umd.edu}
}
\maketitle
\def\thefootnote{*}\footnotetext{These authors contributed equally to this work}\def\thefootnote{\arabic{footnote}}
\begin{abstract}
We present a novel approach for action recognition in UAV videos. 
Our formulation is designed to handle occlusion and viewpoint changes caused by the movement of a UAV. We use the concept of mutual information to compute and align the regions corresponding to human action or motion in the temporal domain. 
This enables our recognition model to learn from the key features associated with the motion. We also propose a novel frame sampling method that uses  joint mutual information to acquire the most informative frame sequence 
in UAV videos. We have integrated our approach with X3D and evaluated the performance on multiple datasets. In practice, we achieve 18.9\% improvement in Top-1 accuracy over current state-of-the-art methods on UAV-Human\cite{li2021uav}, 7.3\% improvement on Drone-Action\cite{perera2019drone}, and 7.16\% improvement on NEC Drones\cite{choi2020unsupervised}. The code is available at \href{https://github.com/Ricky-Xian/MITFAS}{https://github.com/Ricky-Xian/MITFAS}
\end{abstract}
\begin{figure}[t]
    \centering
    \includegraphics[width=\linewidth]{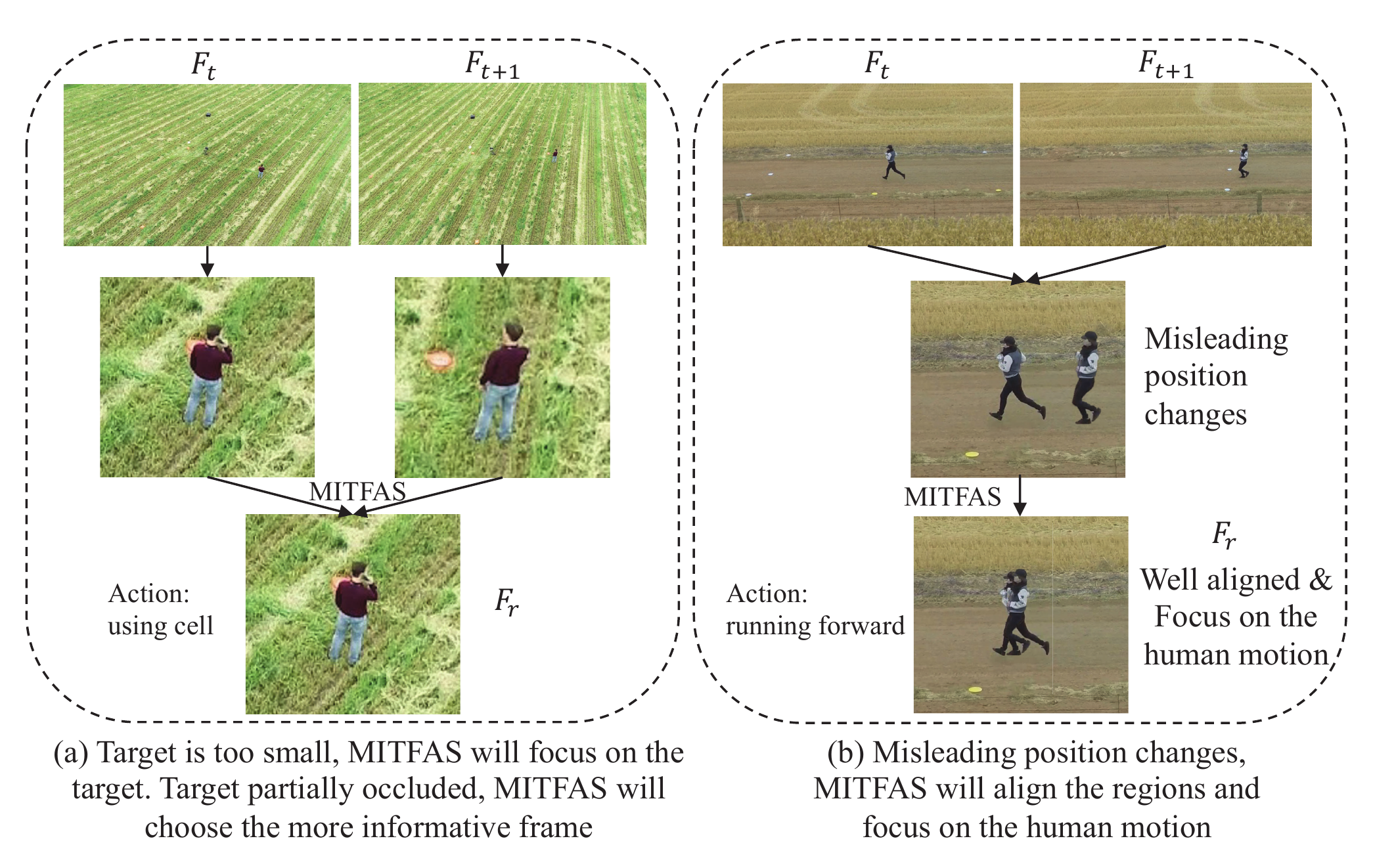}
    \vspace{-6mm}
    \caption{$F_t$ and $F_{t+1}$ are two frames at time $t$ and $t+1$, respectively, from the same UAV video. The human actor in the two frames occupies less than 10\% of the pixels due to the high camera altitude (top images). (a) MITFAS will focus on the regions corresponding to salient motions and use mutual information to find the more informative frame. (b) Because of the UAV's motion, the position of the human actor in $F_{t+1}$ appears to be relatively behind compared to $F_t$. Our algorithm (MITFAS) computes and aligns these regions so that the recognition model will infer more from the human motions. As shown in the right image, the main body of the human actor in two frames overlaps after feature alignment.  
    }
    \label{fig:intro}
    \vspace{-2mm}
\end{figure}

    
    
\section{Introduction}
Unmanned aerial vehicles (UAVs) are increasingly used for different applications, including search and rescue, agriculture, security, construction and aerial surveillance.  This results in many challenging perception problems related to detection, tracking, re-identification, and recognition.
In particular, action recognition using UAV videos is an important problem. While deep learning based methods\cite{feichtenhofer2020x3d,carreira2017quo} have achieved good performance for video action recognition on ground camera videos\cite{carreira2017quo,Monfort2020MomentsIT}, there are many challenges with respect to using them on aerial videos. 


Compared to ground camera videos, the human actors in UAV videos appear rather small due to high camera altitude (see Figure~\ref{fig:intro}). A wider area of the background occupies most of the pixels in the video frame, and only a small fraction (e.g., less than 10\%) corresponds to a human action. Since these videos are captured from a moving (or dynamic) UAV, 
the position and orientation of the human actor may change considerably between the frames. This can result in making the model infer more from the background changes, as opposed to action information, during training. The motion of the UAV camera can also result in blurry frames and some techniques have been proposed to handle them~\cite{li2021uav,zhao2021janusnet,divya2022far}. 


It is harder to collect and annotate UAV videos.  Overall, there are fewer and smaller UAV video datasets, as compared to ground video datasets. 
Additionally, because of continuous changes in the altitude and the camera angle, videos captured using UAVs tend to be more diversified and have unique viewpoints. Some parts of the human actor may be occluded, and not all parts of the human body that contribute to the action can be seen from the camera. Hence, some of the frames in the video are less informative, and this reduces the overall accuracy~\cite{zhi2021mgsampler,Wu2019AdaFrameAF,Ren2020BestFS,Gowda2021SMARTFS}. 



\begin{figure*}[tp]
    \centering
    \includegraphics[width=\linewidth]{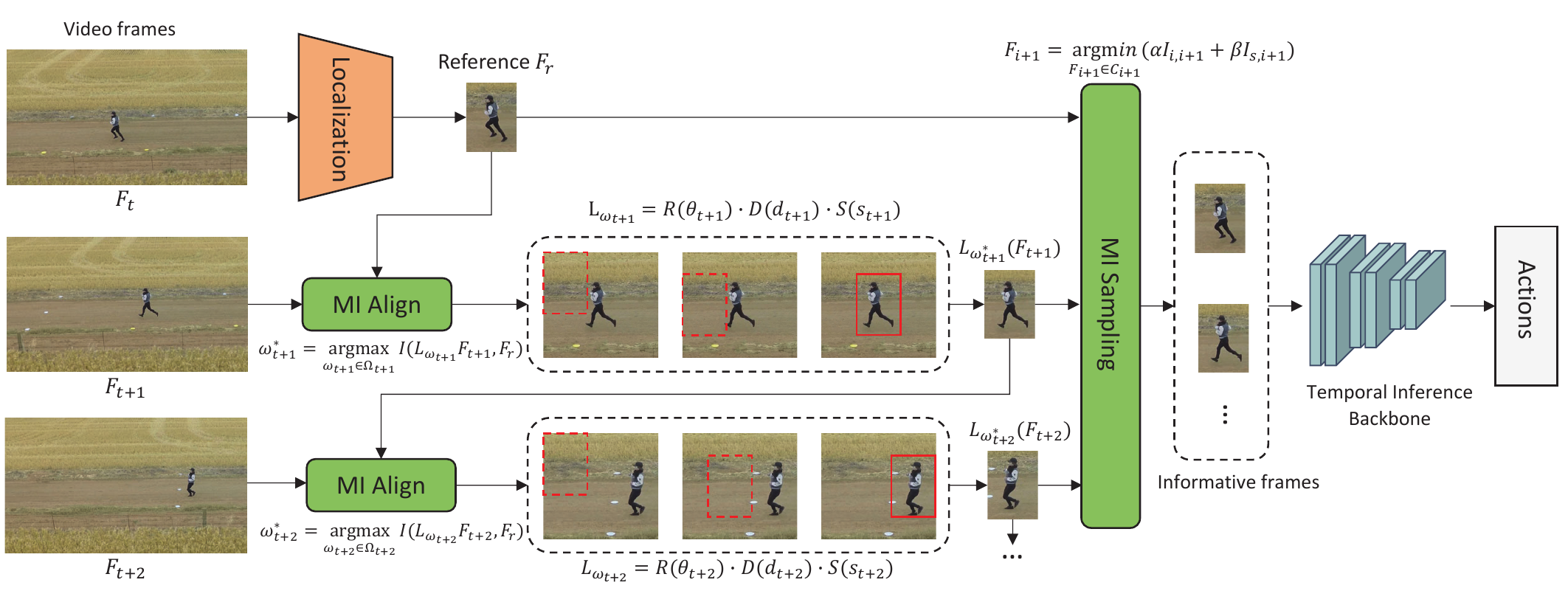}
    \vspace{-7mm}
    \caption{Given a starting frame $F_t$ in a UAV video, we use a localization network to localize the human action and crop the region containing the human motion as the reference image $F_r$. At time $t+1$, we use our feature alignment algorithm 
    to estimate the optimal operation parameter $\omega_{t+1}^*$ 
    and find a region in $L_{\omega_{t+1}^*}(F_{t+1}) \subset F_{t+1}$ that the mutual information between $L_{\omega_{t+1}^*}(F_{t+1})$ and the reference image $F_r$ is maximized. 
    Next, we use $L_{\omega_{t+1}^*}(F_{t+1})$ as the new reference image to find the optimal parameter $\omega_{t+2}^*$ at time $t+2$ and repeat for subsequent frames. Then, we use the criterion illustrated in Section.~\ref{subsec:sampling} Eq .~\ref{eq:sample2} to find a sequence of the most distinctive and informative frames. We use a temporal inference backbone network (e.g., X3D\cite{feichtenhofer2020x3d}) to generate the predicted action label from the spatial-temporal features associated to the sampled frame sequence.}
    \label{fig:overview}
    \vspace{-2mm}
\end{figure*}

\paragraph{Main Contribution:}
We present a novel approach for video action recognition in UAV videos with dynamic backgrounds and moving cameras. 
We take advantage of the mutual information to obtain and align the useful features corresponding to the human actor in the temporal domain. Our alignment method is used to identify the region of the human action, and find the most similar features in the video sequence. As a result, our learning-based recognition model is able to focus more on the human action, rather than the background  regions. Due to the varying viewpoints generated by the movement of a UAV camera, not all the human body parts that contribute to the action are visible.  We present a  novel frame sampling method based on joint mutual information  for dynamic UAV videos, which can compute the most informative and distinctive frame sequence for training aerial action recognition models. We have integrated our temporal feature alignment and frame sampling methods with X3D\cite{feichtenhofer2020x3d} and use them for aerial action recognition (as shown in Figure~\ref{fig:overview}). The novel components of our work include: 
\begin{enumerate}
    \item We use mutual information as a criterion to obtain and align the features at the same time. Our method takes the movement of the UAV into account and estimates the overlapping features by maximizing the mutual information. Given a reference image frame, our approach finds the most similar features in the subsequent frames.
    \item We present a new frame sampling method for UAV videos. Our approach is designed to compute the most informative frame sequence in the video such that all the frames are mostly different from each other. We combine mutual information and joint mutual information to extract the frame.
    Our method is flexible and can deal with different variations in the video sequences. Extensive experiments show our sampling method overperforms peers.   
\end{enumerate}
We test our method on 3 public UAV video datasets. We achieve 20.2\% improvement over the baseline method and  18.9\% improvement over current state-of-the-art method on UAV-Human~\cite{li2021uav}. Our method improves the top-1 accuracy on Drone Action ~\cite{perera2019drone} by 16.6\% over the baseline method and 7.3\% over the current state-of-the-art methods. On NEC Drones~\cite{choi2020unsupervised}, our method get 78.62\% top-1 accuracy, which is 7.18\% higher than the current state-of-the-art and 12.47\% over baseline model by using $1/2$ input frame size. 
\label{sec:intro}

\section{Related Work}
\label{sec:related}
\subsection{Temporal Feature Alignment}
Temporal alignment-based methods have been extensively studied in various video tasks. For instance, Cao et al. \cite{cao2020few} proposed an ordered temporal alignment algorithm specifically designed for few-shot video classification. Similarly, Lu et al. \cite{lu2019indices} introduced an index-guided framework that utilizes indices to guide pooling and up-sampling operations, while Huang et al. \cite{huang2021fapn} presented a method for learning transformation offsets of pixels to align up-sampled feature maps. Additionally, Huang et al. \cite{huang2021alignseg} proposed an aligned feature aggregation algorithm for aligning features of multiple resolutions, and Liu et al. \cite{liu2022temporal} explored feature alignment in the context of multi-frame human pose estimation.

However, it is important to note that most of these methods rely on skeleton information and employ learning-based modules to align important joint points. Despite their success in ground-based videos, these methods may not be as effective in aerial videos due to two primary reasons. Firstly, there is a scarcity of labeled datasets for aerial videos compared to ground videos, making it more challenging to train these learning modules and difficult to transfer them to unseen domains. Secondly, the high flying altitude and moving camera in aerial videos make it challenging to accurately identify important body joint points. To address these limitations, our proposed method takes a different approach by avoiding complex training procedures and not relying on skeleton information.

\subsection{Similarity Measurement}

Various similarity measures have been proposed for comparing image patches. However, in the case of UAV videos with small actor resolution and moving cameras, conventional metrics like Euclidean distance~\cite{zhi2021mgsampler} are affected by background changes and shaking frames. Cosine similarity~\cite{hoe2021one}, although used for high-dimensional data, neglects pixel value magnitudes. Peak Signal-to-Noise Ratio (PSNR)~\cite{hore2010image} focuses on pixel-level comparisons but is sensitive to dominant background changes. Structural Similarity Index Measure (SSIM)~\cite{setiadi2021psnr} evaluates luminance, contrast, and structure but is susceptible to structural variations like rotations and shifts commonly found in aerial videos.

Mutual information is used as a similarity measure between images by~\cite{Viola1995AlignmentBM,Maes1997MultimodalityIR}.  As a similarity measure, mutual information has been widely used in the medical imaging domain~\cite{Pluim2003MutualinformationbasedRO,Klein2007EvaluationOO}. Liu et al.~\cite{liu2022temporal} have explored the possibility of using mutual information for person pose estimation tasks. Ji et al.~\cite{Ji2018InvariantID} have proposed an unsupervised image clustering and segmentation method by maximizing the mutual information between spatial region pairs.
Inspired by the success of mutual information for image processing, we use this concept for temporal feature alignment and frame sampling.  Compared to other similarity measures, mutual information measures the statistical dependence or information redundancy between two images using pixel value distributions, which makes it more robust.

\subsection{Video Recognition for Aerial Videos}
Aerial video action recognition is a challenging task, especially when the camera is moving. The performance of action recognition on ground-camera video datasets has increased as a result of recent advancements in deep learning techniques. However, we don't get a similar level of accuracy on videos captured using UAV cameras~\cite{nguyen2022state}. For aerial video, \cite{geraldes2019uav},\cite{mliki2020human},\cite{mishra2020drone},\cite{mou2020event},\cite{barbed2020fine},\cite{gammulle2019predicting},\cite{mou2020event} apply 2D CNNs (e.g., ResNet, MobileNet) as the backbones to perform single-frame classification and combine the outputs of all frames in the video for recognition. \cite{barekatain2017okutama},\cite{perera2019drone},\cite{perera2020multiviewpoint} leverage two-stream CNNs to utilize attributes from the human motion and the appearance. \cite{choi2020unsupervised},\cite{demir2021tinyvirat},\cite{li2021uav},\cite{mou2020event},\cite{sultani2021human} use I3D network~\cite{carreira2017quo} to learn from spatial-temporal features from human actors and surroundings. To better focus on the target actor in the video, \cite{divya2022far,kothandaraman2023frequency} have proposed an attention mechanism with Fourier transform for better feature extraction. AZTR~\cite{wang2023aztr} proposes a general framework leveraging CNNs and attention mechanisms for aerial action recognition on both edge devices and decent GPUs.  Our feature alignment and sampling method could also be combined with these action recognition methods to improve their accuracy.

Given a video captured from a UAV,  classic feature representation algorithms for aerial video action recognition are limited by the small size of the human actors in aerial videos. 
Sometimes, these approaches improperly identify the camera's motion as a feature~\cite{washington2021activity, mi2020moving}.  \cite{jain2013better} have proposed 2D affine motion models to approximate the camera motion between the adjacent frames. \cite{jiang2012trajectory} have proposed a method where the motion patterns of dense trajectories are clustered to characterize foreground-foreground or foreground-background relationships. Inspired by prior works, our method aligns the human-centered views that are transformed from UAV videos to learn from key features corresponding to the parts of the human body that contribute most to the actions.


\section{Video Recognition using Mutual Information}


We present a mutual information-based method for action recognition on UAV videos with moving cameras and dynamic backgrounds. Our method takes the characteristics of the UAV videos into consideration and uses mutual information as the criterion to compute and align the regions that existing salient motions in the video.
We use joint mutual information to sample the frame sequences that convey most information about human action. Table.~\ref{tab:notation} highlights the notation and symbols used in this section. {We provide the foundational understanding of mutual information in Appendix~\ref{sec:imple}, urging readers to review it beforehand for an enhanced grasp of the paper.
\begin{table}
\centering
\resizebox{0.8\columnwidth}{!}{
\begin{tabular}{c c c}
\toprule
Notation  & Term   \\
\midrule
I()  &  Mutual information, joint mutual information \\
H() &  Entropy, joint entropy \\
p & Probability mass function\\
h & Joint histogram\\
F  &  Frame sequence in the video\\
L  &  Operations to get aligned region \\
R & Rotation matrix\\
D & Translation matrix\\
S & Scaling operation\\
$\omega_t$ & Operations parameters\\
M & Mapping function from frames to features\\
C & Candidate pool for frame sampling\\
\bottomrule
\end{tabular}
}
\caption{Notation and symbols used in the paper.}
\vspace{-9pt}
\label{tab:notation}
\end{table}

\subsection{Temporal Feature Alignment}\label{sec:tfa}

In this section, we describe our approach that uses mutual information (illustrated in Appendix.~\ref{sec:mi}) to obtain and align the features that correspond to salient motions in the temporal domain. In UAV videos, human actors appear significantly small in aerial data, and most pixels in the frame belong to the background. Therefore, we have redundant information about the background in the video that may decrease the performance of our learning model. Moreover, the position of the human actor may change considerably between adjacent frames, which makes the recognition model infer more from the pixels corresponding to redundant background information than the human body movements.
Thus, our objective is to find the region that contains dominant information about the action for each frame in the video and the pixels related to the human actors are well matched.

Let's assume that all the images have the same 2D image coordinate with the origin positioned in the top left corner, with the $x$ axis along the rows and $y$ axis along the columns. Given a video $V$, which corresponds to a sequence of raw frames at different times, $V=\{\cup F_t, t\in N\}$. We generate the reference image $F_r$ that is transformed from a region in the raw frame $F_t$. The reference image $F_r$ is a human centred image that mainly contains salient actions of the human actor.

 To compute $F_r$, suppose $\Omega_t$ contains all feasible operation parameters $\omega_t$, such that for $\omega_t \in \Omega_t$, we can generate a region from $F_t$ using an operation $L_{\omega_t}$. We can consider $L_{\omega_t}$ as a transformation from 2D raw frame coordinates of $F_t$ to the 2D reference frame coordinates corresponding to $F_r$, followed by scaling to the same size of $F_r$. Thus, $L_{\omega_t}$ consists of rotation operation $R(\theta_t)$, translation operation $D(d_t)$ and scaling operation $S(s_t)$, where $\omega_t=(\theta_t,d_t,s_t) \in \Omega_t$: 

\begin{equation}
    L_{\omega_t} = R(\theta_t)\cdot D(d_t) \cdot S(s_t)
\end{equation}

\noindent Our objective is to find $\omega_t^*\in \Omega_t$ for every $t$ such that:
\begin{equation}
    \omega_t^* = \arg\max_{\omega_t\in \Omega_t} I(L_{\omega_t}(F_t);F_r),
\end{equation}
where
\begin{equation}
    I(L_{\omega_t}(F_t);F_r) = H(L_{\omega_t}(F_t)) + H(F_r)-H(L_{\omega_t}(F_t), F_r).
\end{equation}
We use this equation to compute the optimal parameter $\omega_t^*$, so as to compute the target region in $F_t$ that is aligned with $F_r$.

 We need to calculate the mutual information between two images $L_{\omega_t}(F_t)$ and $F_r$. There is no exact mathematical model known to precisely calculate the actual probability distributions related to each image. In general, marginal and joint histograms are  used~\cite{Viola1995AlignmentBM} to approximate the respective distributions. Let $v_{\omega_t}(p)$ denote the value of the pixel at position $p$ in $L_{\omega_t}(F_t)$ and $z_{\omega_t}(p)$ the intensity of the corresponding pixel in $F_r$. The joint histogram $h_{\omega_t}(v,z)$ can be computed by binning the values of the pixel pairs $(v_{\omega_t}(p),z_{\omega_t}(p))$ for all possible $p$. We conduct ablation experiments on the impact of bin numbers that are used to generate histogram in Section.~\ref{subsec:ablation}. Then, the marginal probability distribution $p_{V \omega_t}(v)$,$p_{Z \omega_t}(z)$ and joint probability distribution $p_{VZ \omega_t}(v,z)$ of $v$ and $z$ can be obtained by normalizing the joint histogram $h_{\omega_t}(v,z)$:
\begin{equation}
\begin{split}
    p_{VZ\omega_t}(v,z) &= \frac{h_{\omega_t}(v,z)}{\sum_{v,z}h_{\omega_t}(v,z)},\\
    p_{V\omega_t}(v) &= \sum_{z} p_{VZ\omega_t}(v,z),\\
    p_{Z\omega_t}(z) &= \sum_{v} p_{VZ\omega_t}(v,z).\\
\end{split}
\label{eq:mi_pd}
\end{equation}
The mutual information can be calculated as:
\begin{equation}
    I(L_{\omega_t}(F_t);F_r)=\sum_{v,z}p_{VZ\omega_t}(v,z)\log\frac{p_{VZ\omega_t}(v,z)}{p_{V\omega_t}(v)p_{Z\omega_t}(z)}
\end{equation}
Mutual information is computed using histograms of low-level pixel values on both target and reference patches, which is similar to the mean shift tracking. However, our method uses histograms to approximate the joint probability distribution and measure the inherent statistical dependence between target and reference patch. Also, it can be applied at the feature level. We use a feature extractor to get the features for both $F_t$ and $F_r$. Suppose the mapping function between the RGB images to the features is $M$, the features extracted from $F_t$ and $F_r$ are $M(F_t)$ and $M(F_r)$. Our objective reduces to finding a subset $M_s(F_t) \subset M(t)$ such that
\begin{equation}
    M_s(F_t)^* = \arg\max_{M_s(F_t)\subset M(F_t)} I(M_s(F_t);M(F_r)),
\end{equation}
where
\begin{equation}
\begin{split}
         I(M_s(F_t);M(F_r)) = &H(M_s(F_t)) + H(M(F_t)) \\
         &- H(M_s(F_t),M(F_t)).
\end{split}
\end{equation}

\begin{figure}[tp]
    \centering
    \includegraphics[width=\linewidth]{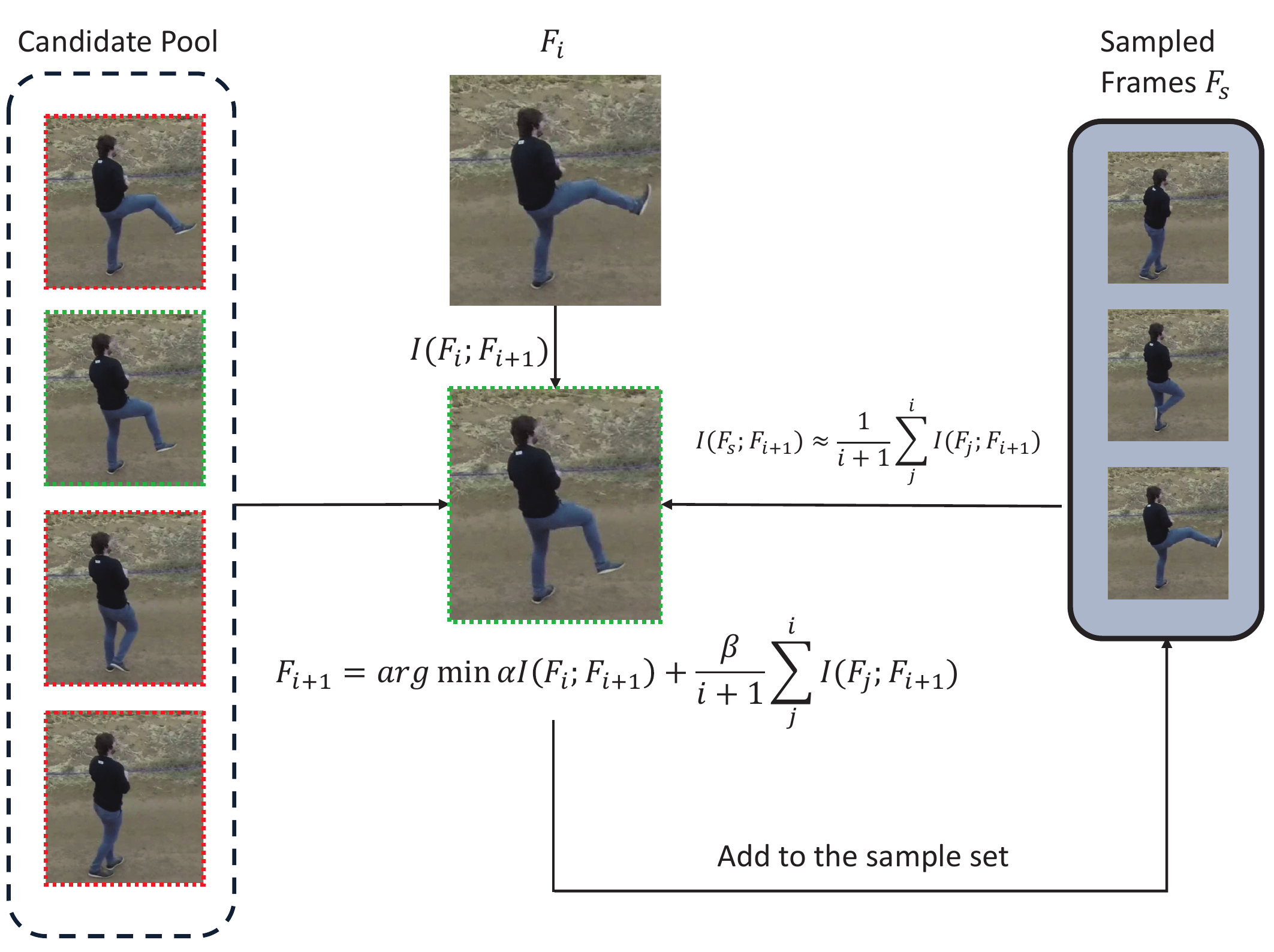}
    \vspace{-7mm}
    \caption{We sample the $i+1th$ frame $F_{i+1}$ from the candidate pool by choosing the frame that is not only least similar to the previous frame but also all the previously sampled frames.}
    \label{fig:sampling}
    \vspace{-2mm}
\end{figure}

 \begin{table*}[!t]
\centering
\resizebox{0.95\textwidth}{!}{
\begin{tabular}{c c c c c c}
\toprule
Method & Backbone & Frames Number & Input Size & Initialization & Top-1 Acc. (\%) $\uparrow$   \\
\midrule 
X3D-M \cite{feichtenhofer2020x3d}& - & $16$ & $224\times224$ & None & $27.0$ \\
X3D-L \cite{feichtenhofer2020x3d}& - & $16$ & $224\times224$ & None & $27.6$ \\
FAR \cite{divya2022far} & X3D-M & $16$ & $224\times 224$ & None & $27.6$ \\
\textbf{Ours (MITFAS)} & X3D-M & $16$ & $224\times 224$ & None & \textbf{40.2} \\
\midrule 
FAR \cite{divya2022far} & X3D-M & $8$ & $540\times 540$ & None & $28.8$ \\
\textbf{Ours (MITFAS)} & X3D-M & $8$ & $540\times 540$ & None & \textbf{38.4} \\
\midrule 
I3D \cite{carreira2017quo}& ResNet-101 & $8$ & $540\times960$ & Kinetics & $21.1$ \\
FNet \cite{lee2021fnet} & I3D & $8$ & $540\times960$ & Kinetics & $24.3$ \\
FAR \cite{divya2022far} & I3D & $8$ & $540\times960$ & Kinetics & $29.2$ \\
FAR \cite{divya2022far} & X3D-M & $8$ & $620\times620$ & Kinetics & $39.1$ \\
\textbf{Ours (MITFAS)} & X3D-M & $8$ & $620\times 620$ & Kinetics & \textbf{46.6} \\
\midrule 
X3D-M  \cite{feichtenhofer2020x3d}& - & $16$ & $224\times224$ & Kinetics & $30.6$ \\
MViT  \cite{Fan2021MultiscaleVT}& - & $16$ & $224\times224$ & Kinetics & $24.3$ \\
FAR \cite{divya2022far} & X3D-M & $16$ & $224 \times 224$ & Kinetics & $31.9$ \\
\textbf{Ours (MITFAS)} & X3D-M & $16$ & $224\times 224$ & Kinetics & \textbf{50.8} \\
\bottomrule 
\end{tabular}
}
\vspace{-7pt}
\caption{Benchmarking UAV Human and comparisons with prior arts.. For $224\times 224$ resolution and 16 frames input, when training from scratch, our approach achieves a $13.2\%$ improvement over the baseline X3D-M and $12.6\%$ over the current state-of-the-art FAR. For $520\times 520$ resolution and 8 frames input, MITFAS overperforms the current state-of-the-art FAR by $9.6\%$ when training from scratch. For $224\times 224$ resolution and 16 frames input, when initializing with Kinetics pre-trained weights, MITFAS improves the top-1 accuracy over baseline by $20.2\%$ and over SOTA method by $18.9\%$. For resolution over $620\times 620$ and  8 frames input, when initializing with Kinetics pretrained weights, MITFAS overperforms the current state-of-the-art FAR by $7.5\%$. Our method obtains better performance in all settings, which illustrates the effectiveness of our proposed MITFAS.}
\vspace{-7pt}
\label{tab:resolution}
\end{table*}
\subsection{Mutual Information Sampling}\label{subsec:sampling}
Because of the high camera altitude, many parts of the human body are not visible. Some parts of the human body that result in the action may be occluded by some other parts that do not contribute to the action. Also, there are lots of "duplicated" frames because of the high frame rate, which essentially contains redundant information. Therefore, not all the video frames are useful for the training, and using some of them may even decrease the overall accuracy. To solve this issue, we present a novel frame sampling method using a combination of mutual information and joint mutual information to find the frame sequences that contain more information about action changes in the UAV videos. 

The main idea behind our method is to find out more informative frame sequences in the video given a start frame. Consider a video as a sequence of frames across time. Suppose we have already sampled $i$ frames and our goal is to find the $i+1$th frame $F_{i+1}$ in the candidate pool $C_{i+1}$ where $C_{i+1}$ consists of all the possible frames that we could choose for $F_{i+1}$. Let $F_s = \{\cup F_0, F_1, F_2 \cdots F_i\}$ denote the set that contains all the sampled frames. Our approach is to choose $F_{i+1}$ that is the most distinctive as compared with $F_i$ as well as the set of all previously sampled frames so that it provides more unseen features for the recognition model training, see Figure~\ref{fig:sampling}:
\begin{equation}\label{eq:sampling}
    F_{i+1} = \arg \min_{F_{i+1}\in C_{i+1}} \alpha I(F_i;F_{i+1}) + \beta I(F_s;F_{i+1}).
\end{equation}
The first term is used to minimize the mutual information between the current frame and the previous frame. It tends to sample adjacent frames that are least similar so that the newly sampled frame will contain more information for training. 
The second term is used to minimize the joint mutual information with all the sampled frames, which could decrease the information redundancy over the whole sampling sequence. We can decompose it using the chain rule of joint mutual information:
\begin{equation}\label{eq:sampling2}
\begin{split}
         I(F_s;F_{i+1}) &= I(F_0,F_1,F_2 \cdots F_i;F_{i+1})\\
         &= \sum_{j=0}^i I(F_j;F_{i+1}|F_{j-1}F_{j-1}\cdots F_{0})
\end{split}
\end{equation}
In practice, the conditional mutual information is hard to compute as the conditional probability distribution is hard to calculate. However, to make the problem more tractable, we use the low-dimensional approximation to estimate the joint mutual information between $F_{i+1}$ and $F_s$ \cite{Gao2017EstimatingMI,JMLR:v13:brown12a}.
\begin{equation}
\begin{split}
         I(F_s;F_{i+1})&\approx \frac{1}{i+1}\sum_{j=0}^i I(F_j;F_{i+1})\\
\end{split}
\end{equation}
So the overall expression becomes:
\begin{equation}\label{eq:sample2}
    F_{i+1} = \arg \min_{F_{i+1}\in C_{i+1}} \alpha I(F_i;F_{i+1}) + \frac{\beta}{i+1} \sum_{j=0}^i I(F_j;F_{i+1})
\end{equation}
Here, we add weights $\alpha$,$\beta$ to the two terms in Eq.~\ref{eq:sample2} to adjust to different scenarios. We analyze the behavior of $\alpha$ and $\beta$ in the Appendix ~\ref{app: mi_sampling}.

\subsection{MITFAS: Aerial Video Recognition} \label{subsec:MITFAS}
In this section, we present our overall method for aerial video recognition (see Fig.~\ref{fig:overview}). We use temporal feature alignment and frame sampling and combine them with a temporal inference backbone network (e.g, X3D\cite{feichtenhofer2020x3d}) to disentangle the human actor from superfluous backgrounds and learn from key features associated with the human motions.


In our benchmarks, most of the videos available are captured on a UAV camera with anti-shake technology which could stabilize the camera and reduce the camera vibration, we assume no rotation is needed, i.e., $R(\theta_t)=Identity$. For general videos, $R(\theta_t)$ is the rotation matrix represented and computed as a 2D transformation:
\begin{equation}
    R(\theta_t) = \begin{bmatrix}
        \cos\theta_t & -\sin\theta_t\\
        \sin\theta_t & \cos\theta_t
    \end{bmatrix}
\end{equation}

We localize the human actor at the start frame and enlarge the region by about 10\% of its height to obtain the reference $F_r$ ~\cite{hasan2021generalizable}. We conduct ablation studies on the size of $F_r$ in Section.~\ref{subsec:ablation}. Considering the human actor may perform actions that have large vertical changes like stretching arms, we add 15\% height as the margin on the top of $F_r$ to ensure all the information about the action are included and crop the region as our final reference image. Therefore, we enlarge the region by 25\% vertically and 10\% horizontally to get $F_r$.

We use the sliding window strategy with scalable window sizes to find the aligned regions or features in all the frames. To make the process more efficient, we do not apply sliding window search over the entire frame. Instead, once we compute $\omega_t^*$ at time $t$, we use the same operation at $t+1$ to obtain the region $L_{\omega_t^*}(F_{t+1})$. We expand  $L_{\omega_t^*}(F_{t+1})$ by 25\% as the searching area at $t+1$. In this way, we could significantly decrease the overall mutual information computations by only searching in the searching area which is a subset of $F_{t+1}$. In order to improve the reliability, we occasionally re-perform localization to update the searching area. More ablation studies on the impact of searching area size is given in the supplementary.

Once all the $\omega_t^*$ are found for all time $t$,  well-aligned frames are obtained by the transformation. We use our frame sampling method illustrated in Section.~\ref{subsec:sampling} to generate a sequence of 8 or 16 frames for model training. We will randomly pick a start frame as $F_0$, denoting the index of $F_0$ in the sequence as $k_0$. To maintain the randomness in our sampling strategy, we set a randomly generated stride $r_1$ when sampling $F_1$. We compute our candidate pool $C_1$ by a set of all the frames that have index greater than $k_0$, but not exceed $k_0 + r_1$. Next, we find the most informative frame in the candidate pool using Eq.~\ref{eq:sample2} and use it as $F_1$. We follow the same strategy to sample all the subsequent frames.

After obtaining all the sampled frames, we use a temporal inference backbone network to extract and learn from spatial-temporal features from the human actions. We employ X3D\cite{feichtenhofer2020x3d} as the backbone in our method for its efficiency and performance on video tasks. However, our method could be combined with any action recognition models for better behavior understandings on UAV videos.

\label{sec:method}

\section{Results}
\label{sec:results}

\begin{table}
\centering
\resizebox{1.0\columnwidth}{!}{
\begin{tabular}{c c c c c}
\toprule
Method & Frames & Input Size & Init. & Top-1   \\
\midrule
HLPF &  All & $1920\times1080$ & None & $64.3$ \\
PCNN &  - & $1920\times1080$ & None & $75.9$\\
X3D-M &  $16$ & $224\times224$ & Kinetics & $83.4$ \\
FAR &  $16$ & $224\times224$ & Kinetics & $92.7$ \\
\textbf{Ours} &  $16$ & $224\times224$ & Kinetics & \textbf{100.0} \\
\bottomrule
\end{tabular}
}
\vspace{-7pt}
\caption{Results on Drone Action. Our method achieves 100\% top-1 accuracy, 16.6\% over the baseline method X3D-M\cite{feichtenhofer2020x3d}, outperforming current state-of-the-art method FAR\cite{divya2022far} by 7.3\% under same configuration. (HLPF \cite{jhuang2013towards}, PCNN \cite{cheron2015p})}
\vspace{-5pt}
\label{tab:droneaction}
\end{table}

\begin{table}
\centering
\resizebox{1.0\columnwidth}{!}{
\begin{tabular}{c c c c c}
\toprule
Method & Frames & Input Size & Init. & Top-1   \\
\midrule
X3D-M &  $8$ & $960\times540$ & Kinetics & $66.1$ \\
FAR &  $8$ & $960\times540$ & Kinetics & $71.4$ \\
\textbf{Ours} &  $8$ & $540\times540$ & Kinetics & \textbf{78.6} \\
\bottomrule
\end{tabular}
}
\vspace{-7pt}
\caption{Results on NEC Drones. Our method shows an improvement of 12.5\% on top-1 accuracy against the baseline X3D-M\cite{feichtenhofer2020x3d}, 7.2\% over current state-of-the-art FAR \cite{divya2022far}.}
\vspace{-7pt}
\label{tab:more_drones}
\end{table}

In this section, we describe our implementation and present the results. We compare the performance with other state-of-the-art video action recognition methods on 3 UAV datasets. The implementation and training details are shown in Appendix.~\ref{sec:imple}.




\subsection{Results on UAV Human}
UAV Human is currently the largest UAV-based human behavior understanding dataset. It contains scenarios captured from both indoor and outdoor environments with different lighting and weather conditions. The videos are captured in dynamic backgrounds with different UAV motions and flying altitudes. It has 155 annotated actions, many of which are hard to distinguish such as squeeze and yawn.

We compare our method against prior state-of-the-art methods on UAV Human. As shown in Table~\ref{tab:resolution}, we implement our method and compare the performance with other state-of-the-art methods in various configurations in terms of the backbone network, frame rates, frame input sizes, and weights initialization. We use X3D-M as the backbone of our method with two different initialization settings. One of them is training from scratch and the other is initialized with Kinetics pretrained weights. 

First, when using the same configuration (frames, input size, initialization), our method outperforms all the prior methods by a large margin. When training from scratch, we achieve a 12.6\% improvement over current state-of-the-art methods. We get an 18.9\% improvement when using Kinetics pretrained weights. This indicates the effectiveness of our method, which reduces the information redundancy and makes the model learn more from the motion changes rather than background variations.




\subsection{Results on NEC Drone}
NEC Drone is an indoor dataset that contains 5,250 videos with 16 actions performed by 19 actors. The videos are captured using a UAV flying at a low altitude on a basketball court. Compare to UAV Human, NEC Drone has more consistent lighting conditions while bringing more noises caused by light reflections.

We present the results on NEC Drone in Table \ref{tab:more_drones}. We obtain a Top-1 accuracy of 78.6\%. We compare our method against the baseline X3D-M and shows an improvement of 12.5\%. Our approach outperforms the current SOTA FAR on NEC Drone by 7.2\%. Note that, the improvement we achieved is obtained with $1/2$ input frame size, which further demonstrates the advantage of our method.
\subsection{Results on Drone Action}
Drone Action is an outdoor video dataset that was captured using a free-flying UAV in low altitude and low speed. It contains 240 videos across 13 human actions performed by 10 human actors. Drone Action is the smallest dataset we used, but it is collected using a free-flying UAV that results in continuous position changes of the human actor. 

As shown in Table \ref{tab:droneaction}, we achieve 100\% Top-1 accuracy which outperforms current SOTA by 7.3\% under the same configuration, which further illustrates the benefits of our proposed MITFAS.

\begin{table}
\centering
\resizebox{0.95\columnwidth}{!}{
\begin{tabular}{c c | c c}
\toprule
Sampling Method & Top-1 & Sampling Method & Top-1  \\
\midrule
Random & $23.8$ & TFA + Random & $39.8$ \\
Uniform & $25.8$ & TFA + Uniform & $42.2$ \\
MG Sampler & $28.1$ & TFA + MG Sampler & $45.5$ \\
MIS & \textbf{28.7} & TFA + MIS  & \textbf{46.2} \\
\bottomrule
\end{tabular}
}
\vspace{-7pt}
\caption{Temporal Feature Alignment (TFA) and Mutual Information Sampling (MIS) ablation studies on UAV-Human-Subset. The baseline is vanilla X3D with random~\cite{fischler1981random} and uniform sampling~\cite{krizhevsky2017imagenet}, and we add our methods TFA and MIS step by step. From our experiments, TFA boost the accuracy by 16-17.5\%.  MIS outperforms the random sampling, uniform sampling, and MG Sampler\cite{zhi2021mgsampler}.}
\vspace{-5pt}
\label{tab:sampling}
\end{table}
\begin{table}
\centering
\resizebox{0.95\columnwidth}{!}{
\begin{tabular}{c c c}
\toprule
Method  & UAV-Human & Drone Action  \\
\midrule
Bounding box tracking~\cite{demir2021tinyvirat} & $47.4$ & $95.9$\\
Spatial-temporal action detection ~\cite{liu2020argus}& $47.9$ & $95.9$\\
\textbf{TFA(ours)} & \textbf{50.8} & \textbf{100.0}\\
\bottomrule
\end{tabular}
}
\vspace{-7pt}
\caption{Comparison with other methods~\cite{demir2021tinyvirat,liu2020argus}. }
\vspace{-7pt}
\label{tab:bbox_tracking}
\end{table}
\subsection{Ablation Experiments}

In this subsection, we mainly show the results of ablation experiments to demonstrate the effectiveness of the two components of our approach: Temporal Feature Alignment(TFA) and Mutual Information Sampling(MIS). More ablation studies are given in Appendix. 

We randomly pick 30\% videos for each action label in UAV-Human and conduct the ablation experiments on this UAV-Human subset. We use X3D-M\cite{feichtenhofer2020x3d} as the backbone network. All results are generated by using a sequence of 16 frames with a resolution of 224 × 224.

\begin{table}
\centering
\begin{center}
\resizebox{\columnwidth}{!}{
\begin{tabular}{c c c c c }
\toprule
Sampling Method & Alpha & Beta & \makecell{UAV-Human \\ Top-1}  & \makecell{DroneAction \\ Top-1}  \\
\midrule
X3D + TFA + MIS       & 1.0 &  0.0 & $45.3$ & 94.5\\
X3D + TFA + MIS        & 0.0 &  1.0 & $45.7$ & 95.9\\
X3D + TFA + MIS        & 1.0 &  0.5 & $45.5$ & 97.2\\
X3D + TFA + MIS        & 1.0 &  1.0 & \textbf{46.2} & \textbf{100}\\
\bottomrule
\end{tabular}
}
\vspace{-7pt}
\caption{Mutual Information Sampling (MIS) ablation studies on UAV-Human-subset and Drone Action. The baseline is vanilla X3D with TFA, we test the MITFAS Sampling in terms of  two hyperparameters for mutual information and joint mutual information, $\alpha$ and $\beta$ respectively. From our experiments, MITFAS obtains the best accuracy when $\alpha=1.0$ and $\beta=1.0$.  }
\vspace{-5pt}
\label{tab:sampling_ablation}
\end{center}
\end{table}
\begin{table}
\centering
\begin{tabular}{c c }
\toprule
Similarity Measure  & Top-1 Acc  \\
\midrule
Euclidean Distance & 42.1  \\
Cosine Similarity &39.5  \\
Peak Signal-to-Noise Ratio & 43.4 \\
Structural Similarity Index Measure & 44.8  \\
\textbf{Mutual Information} & \textbf{46.2} \\

\bottomrule
\end{tabular}
\vspace{-7pt}
\caption{Comparison with other similarity measures on UAV-Human Subset. Compared to other similarity measures, mutual information achieves the best accuracy.}
\vspace{-8pt}
\label{tab:similarity}
\end{table}
\textbf{Effectiveness of Temporal Feature Alignment} For Temporal Feature Alignment (TFA), our objective is to solve the small resolution corresponding to the human actor and viewpoint changes in the UAV videos. 
Our TFA finds and aligns the region that contains dominant information about the action for each frame in the video. As shown in Table.~\ref{tab:sampling}, our TFA improves the top-1 accuracy by 16 - 17.5\% when it is integrated with X3D and different sampling methods.

We also compare our TFA with other methods in Table~\ref{tab:bbox_tracking}. The bounding box tracking method~\cite{demir2021tinyvirat} applies the person detector for foreground patch detection on all the temporal frames and then extracts the foreground patch based on the bounding boxes. Standard spatial-temporal action detection pipeline~\cite{liu2020argus} integrates the detection and tracking algorithms to generate the proposal for the feature extraction. The results are generated using X3D and uniform sampling with the same configurations. As shown in Table.~\ref{tab:bbox_tracking}, our method improves the top-1 accuracy over other two methods by 2.9\% on UAV-Human and 4.1\% on Drone Action. Such improvement is attributed to our proposed TFA can not only extract the foreground patches but also align all the patches so that the main body of the human actor is well-matched in the temporal domain. Unlike the other two methods, our method does not align the bounding boxes. The bounding boxes are only used to locate the human actor. As illustrated in Section~\ref{sec:tfa}, the alignment in our method is performed at the pixel-level, ensuring that the generated frame closely resembles the preceding one. Therefore, the model could focus on the pixels corresponding to the parts of the human body that contribute most to the actions during training. Moreover, our method does not require any training procedures and could be utilized in any scenarios without domain issues.

\textbf{Effectiveness of Mutual Information Sampling} For Mutual Information Sampling (MIS), our goal is to sample the informative frames that better represent the video for the action recognition methods. We compare it with three other sampling methods. First, we compare with two baseline methods:  (1) Random sampling~\cite{fischler1981random} where frames are randomly picked (2) Uniform sampling~\cite{krizhevsky2017imagenet} where frames are sampled uniformly given a randomly generated start and end point. Then, we compare with the current state-of-the-art MG Sampler \cite{zhi2021mgsampler} which uses an adaptive sampling strategy based on temporal consistency between adjacent frames. As shown in Table.~\ref{tab:sampling}, compared with other sampling methods, MIS results in 0.6 - 6.4\% improvement in Top-1 accuracy for UAV videos, which demonstrates the effectiveness of our proposed method.

\textbf{Hyperparameters}
\label{app: mi_sampling}
We evaluate our Mutual Information Sampling in terms of two hyperparameters for mutual information and joint mutual information, $\alpha$ and $\beta$ in Eq.~ \ref{eq:sample2} respectively. The baseline is vanilla X3D with TFA. As shown in Table~\ref{tab:sampling_ablation}, from our experiments, MITFAS obtains the best accuracy when $\alpha=1.0$ and $\beta=1.0$. This demonstrates that both items in Eq.\ref{eq:sample2} are equally important in discriminating the more informative frames.

\textbf{Comparison of other similarity measures}
We compare the result of using mutual information with other similarity measures in Table.\ref{tab:similarity}. The results demonstrate that mutual information is a better criterion for measuring the similarity between images for UAV videos.

\section{Conclusion, Limitations and Future Work}
We propose a novel approach for video action recognition on UAVs. Our approach is designed to handle the varying and small resolution of the human, large changes in the positions of the human actor between frames, and partially occluded key points of the actions caused by continuous movement of the UAVs.  We present a mutual information-based feature alignment to obtain and align the action features in the temporal domain. Our method is efficient and works well on UAV videos. We also present a novel frame sampling method to find the most informative frames in the video. We compare with prior approaches and demonstrate improvements in Top-1 accuracy on 3 UAV datasets. Our approach has a few limitations. First, we assume there does not exist a long-range spatial relationship between the human actor and the background. Second, we assume the input videos contain only one scripted human agent performing some action. We would like to explore the possibility of extending our method to multi-human or multi-action videos. 

\textbf{Acknowledgement} This work was supported in part by ARO Grants W911NF2110026, W911NF2310046,  W911NF2310352  and Army Cooperative Agreement W911NF2120076
\label{sec:conclusion}

{\small
\bibliographystyle{ieee_fullname}
\bibliography{egbib}
}

\clearpage
\appendix
\label{sec:appendix}

\section{Implementation and Training Details}\label{sec:imple}

\noindent  \textbf{Backbone network architecture:} We use a Cascade Masked RCNN\cite{hasan2021generalizable} that was pretrained on Cityperson\cite{Zhang2017CityPersonsAD} as the localization network to localize the human actor at the start frame. We use X3D-M\cite{feichtenhofer2020x3d} as the temporal inference backbone to give the final predicted label.

\noindent  \textbf{Training details:} All the mutual information calculations are implemented on a high-end desktop CPU (Intel Xeon W-2288 CPU), because current version of CUDA does not support histogram operations on GPUs. Our overall model is trained using NVIDIA GeForce 2080Ti GPUs and NVIDIA RTX A5000 GPUs. We use the same initialization as~\cite{divya2022far}. The initial learning rate is set at $0.1$ for training from scratch and $0.05$ for initializing with Kinetics pretrained weights. Stochastic Gradient Descent (SGD) is used as the optimizer with 0.0005 weight decay and 0.9 momentum. We use cosine/poly annealing for learning rate decay and multi-class cross entropy loss to constrain the final predictions.

\noindent  \textbf{Evaluation:} We evaluate our method and other state-of-the-art methods using Top-1 accuracy score, which is the proportion of the correct predictions to all the samples in the evaluation set.

\begin{table*}[t]
\centering
\resizebox{1.0\textwidth}{!}{
\begin{tabular}{c c | c c | c c | c c}
\toprule
Histogram Bin number & Top-1 & Reference image Size & Top-1 & Sliding Stride & Top-1 & Searching area size &  Top-1 \\
\midrule
32 & $52.3$  & 1.10 $\times$ &  $53.4$ & 5  & $52.7$ & 1.25$\times$ reference size & $52.4$  \\
64 & $52.7$  & 1.25 $\times$ &  $54.0$ & 10 & $53.0$ & 1.50$\times$ reference size & $52.7$  \\
128 & $54.3$ & 1.5 $\times$  &  $53.7$ & 15 & $52.8$ & 2.00$\times$ reference size & $52.5$  \\
256 & $52.7$ & 1.75 $\times$ &  $52.5$ & 20 & $51.1$ & 2.50$\times$ reference size & $52.1$  \\
\bottomrule
\end{tabular}
}
\vspace{-7pt}
\caption{Ablation studies on UAV-Human subset in terms of using different bin numbers to calculate mutual information, reference image size (times of the standard size), using different strides for slipping windows, and searching area size. The best performance is achieved while using 128 histogram bins, reference image size 1.25$\times$ and sliding stride of 10. The size of the searching area does not affect the overall performance of our method. The top-1 accuracy only varies 0.6\% while using different searching area sizes. This demonstrates the robustness of our MITFAS as the larger searching area contains more noises and outliers.}
\vspace{-5pt}
\label{tab:ablation}
\end{table*}

\section{Mutual Information}\label{sec:mi}
Mutual information is a concept in information theory that essentially measures the amount of information given by one variable when observing another variable. It can also be interpreted as the reduction of the uncertainty of one variable given the other. Mutual information is highly correlated with entropy and joint entropy. The mutual information between image pairs $X$ and $Y$ can be equivalently expressed as:
\begin{equation}\label{eq:overallmi}
\begin{split}
     I(X;Y)  &=H(X)+H(Y)-H(X,Y), \\ 
\end{split}
\end{equation}
where $H(X)$ and $H(Y)$ correspond to the entropy of $X$ and $Y$, respectively. The entropy quantifies the complexity of all possible outcomes of $X$ or $Y$. Given $p_X(x)$, $x\in \mathcal{X}$ the probability mass function (PMF) of $X$, the entropy of $X$, $H(X)$ can be calculated as:
\begin{equation}\label{eq:entropy}
 H(X) = -\sum_{x\in \mathcal{X}}p_X(x)\log p_X(x).   
\end{equation}
$H(X,Y)$ is the joint entropy that examines the overall randomness given both $X$ and $Y$:
\begin{equation}\label{eq:jointentropy}
 H(X,Y) = -\sum_{x\in \mathcal{X},y\in\mathcal{Y}}p_{XY}(x,y)\log p_{XY}(x,y),   
\end{equation}
where $p_{XY}(x,y), x\in \mathcal{X},y\in\mathcal{Y}$ is the joint probability distribution of intensities of pixels associated with $X$ and $Y$. The joint entropy $H(X,Y)$ is minimized if and only if there is a one-to-one mapping function $G$ such that $p_X(x)=p_Y(G(x))=p_{XY}(x,G(x))$. It increases when the inherent statistical relationship between $X$ and $Y$ weakens. Therefore, as pixels in $X$ become more distinctive from the counterparts in $Y$, $H(X,Y)$ gets larger and $I(X;Y)$ gets smaller.
Note that, if the image or region pairs $X$ and $Y$ are completely independent from each other, then:
\begin{equation}
\begin{split}
        &H(X,Y) = H(X)+H(Y),\\
        &I(X;Y) = 0.\\
\end{split}
\end{equation}
In our case, we use mutual information to obtain and align the region pairs in the temporal domain of a video. Therefore, $X$ and $Y$ are always correlated and $I(X;Y) \neq 0$. Moreover, as we calculate mutual information using probability distribution of discrete pixels, we use sums instead of integrals in Eq .~\ref{eq:entropy} and \ref{eq:jointentropy}.
We use Eq.~\ref{eq:entropy} and \ref{eq:jointentropy} to express the mutual information on Eq.~\ref{eq:overallmi} using probability distributions. Therefore:
\begin{equation}\label{eq:miprob}
 I(X;Y)=\sum_{x\in\mathcal{X},y\in\mathcal{Y}}p_{XY}(x,y)\log\frac{p_{XY}(x,y)}{p_X(x)p_Y(y)}.
\end{equation}
From the equation above, we can see that the mutual information quantifies the dependence between two random variables by measuring the distance between the real joint distribution $p_{XY}(x,y)$ and the distribution under assumption of complete independence of $p_X(x)p_Y(y)$.

Intuitively, as Viola~\cite{Viola1995AlignmentBM} observes, maximizing the mutual information between two images or regions tends to find the most complex overlapping areas (by maximizing the individual entropy) such that at the same time they explain each other well (by minimizing the joint entropy). 

The joint mutual information is an extension of mutual information. It measures the statistical relationship between a single variable and a set of other variables. Given one image $Y$ and a set of images $X_1,X_2$, the joint mutual information is expressed as:
\begin{equation}
    I(X_1,X_2;Y)= I(X_1;Y)+I(X_2;Y|X_1).
\end{equation}
where $I(X_2;Y|X_1)$ is the conditional mutual information that measures the dependence between $X_2$ and $Y$ when observing $X_1$.

\section{Ablation Experiments}\label{subsec:ablation}
We perform ablation experiments to examine the impact of bin number for histograms to calculate mutual information, reference image size, sliding window stride, searching region and MIS hyperparameters. We randomly pick $30\%$ videos for each action label int UAV-Human and conduct the ablation experiments on this UAV-Human subset. We use X3D-M\cite{feichtenhofer2020x3d} as the temporal inference backbone network. All results are generated by using a sequence of 16 frames with resolution $224 \times 224$. All the results are shown in Table.~\ref{tab:ablation}.

\subsection{Bin Numbers for Histograms} 
We calculate the mutual information between two images by using their probability distributions. However, there is no exact mathematical model known to precisely calculate the actual probability distributions related to each image. 
As we mentioned in Eq.~\ref{eq:mi_pd},  we use marginal and joint histograms to approximate the probability distribution. We obtain the joint histogram by binning pairs of pixel values in the two frames. Therefore, bin number is an important hyper parameter for calculating the mutual information. We explore the effects of the number of bins used to generate the joint histogram on the overall performance. We present the results of using different number of bins in Table \ref{tab:ablation}. It shows that the overall accuracy does not monotonically increase as more bins are used and bins number around 128 will result in the best overall performance. It is reasonable because if the histogram is generated with too few bins, then it can not portray the data very well. If too many bins are used, the histogram will not be able to give a good sense of distribution. Therefore, both large and small bin number will lead to bad approximation of the probability distribution, which makes the mutual information calculation less accurate. Moreover, the memory overhead will exponentially grows as more bins are used because the calculation takes the square times of the bin number. To balance the efficiency and accuracy, we use 128 as the bin number for all the experiments in this paper.

\subsection{Reference Image Size}
Our method needs a reference image without much background information redundancy at the beginning, since we need this inference image as the basis to calculate the mutual information with other frames and eventually obtain a sequence of well aligned regions. However, it is hard to determine how much background information is sufficient enough for aerial recognition as all our videos are captured in the oblique and aerial views with drone cameras. Therefore, we evaluated the impact of different ratio of the background in UAV videos. Let the size of the bounding box generated by the localization network be the standard size. We conduct the experiments on reference images with 4 different sizes (i.e., 1.1 $\times$, 1.25 $\times$, 1.5 $\times$, and 1.75 $\times$ of the standard size). As can see in Table .~\ref{tab:ablation}, when the reference images is 1.25$\times$ of the standard size, we obtain the best performance. Less reference image size makes the model unable to analyze the relationship between the human actor and the surroundings due to less background information. But more background information will bring more noises and outliers, decreasing the overall accuracy.

\subsection{Sliding Window Stride}
After we obtain the reference image, we use it for MI alignment with the subsequent frame. Here we employ sliding window strategy to find the well-aligned regions that correspond to salient motions in the video. While computing the sliding window, the stride is an important element that needs to be considered since it dramatically effects the overall efficiency. Larger stride means less searching time but decreases the accuracy, as shown in Table~\ref{tab:ablation}, stride value at 10 results in the highest accuracy. Therefore, we choose 10 as the sliding window stride for all benchmarks.

\subsection{Searching Region}
As mentioned in Section 3.4, to reduce the overall mutual information computations, once we compute $\omega_t^*$ at time $t$, we use $\omega_t^*$  at $t+1$ to obtain the region $L_{\omega_t^*}(F_{t+1})$. Then, we only search in the searching area which is generated by expanding $L_{\omega_t^*}(F_{t+1})$ by 25\% at $t+1$. Therefore, the size of the searching area is an important hyper parameter for our method. As shown in Table.~\ref{tab:ablation} we conduct experiments with different searching area sizes, $1.25\times, 1.5\times, 2.0\times, 2.5\times$ the size of the $L_{\omega_t^*}(F_{t+1})$, on the UAV-Human subset. Surprisingly, the result shows that the searching area size does not have significant impacts on the overall performance of our method (MITFAS). The top-1 accuracy only varies $0.6\%$ while using different searching area sizes. This demonstrates the robustness of our method, as the larger searching area will contains more noises and outliers. Overall, our MITFAS is robust to outliers and can precisely obtain and align the regions existing salient human motions. Therefore, to reduce the overall training time, we choose the searching area size to be $1.25\times$ the size of $L_{\omega_t^*}(F_{t+1})$ in all other benchmarks in this paper.

\end{document}